\PassOptionsToPackage{unicode}{hyperref}
\PassOptionsToPackage{hyphens}{url}
\PassOptionsToPackage{dvipsnames,svgnames,x11names}{xcolor}
\documentclass[
]{article}

\usepackage{amsmath,amssymb}
\usepackage{iftex}
\ifPDFTeX
  \usepackage[T1]{fontenc}
  \usepackage[utf8]{inputenc}
  \usepackage{textcomp} 
\else 
  \usepackage{unicode-math}
  \defaultfontfeatures{Scale=MatchLowercase}
  \defaultfontfeatures[\rmfamily]{Ligatures=TeX,Scale=1}
\fi
\usepackage{lmodern}
\ifPDFTeX\else  
\fi
\IfFileExists{upquote.sty}{\usepackage{upquote}}{}
\IfFileExists{microtype.sty}{
  \usepackage[]{microtype}
  \UseMicrotypeSet[protrusion]{basicmath} 
}{}
\makeatletter
\@ifundefined{KOMAClassName}{
  \IfFileExists{parskip.sty}{%
    \usepackage{parskip}
  }{
    \setlength{\parindent}{0pt}
    \setlength{\parskip}{6pt plus 2pt minus 1pt}}
}{
  \KOMAoptions{parskip=half}}
\makeatother
\usepackage{xcolor}
\makeatletter
\ifx\paragraph\undefined\else
  \let\oldparagraph\paragraph
  \renewcommand{\paragraph}{
    \@ifstar
      \xxxParagraphStar
      \xxxParagraphNoStar
  }
  \newcommand{\xxxParagraphStar}[1]{\oldparagraph*{#1}\mbox{}}
  \newcommand{\xxxParagraphNoStar}[1]{\oldparagraph{#1}\mbox{}}
\fi
\ifx\subparagraph\undefined\else
  \let\oldsubparagraph\subparagraph
  \renewcommand{\subparagraph}{
    \@ifstar
      \xxxSubParagraphStar
      \xxxSubParagraphNoStar
  }
  \newcommand{\xxxSubParagraphStar}[1]{\oldsubparagraph*{#1}\mbox{}}
  \newcommand{\xxxSubParagraphNoStar}[1]{\oldsubparagraph{#1}\mbox{}}
\fi
\makeatother

\providecommand{\tightlist}{%
  \setlength{\itemsep}{0pt}\setlength{\parskip}{0pt}}\usepackage{longtable,booktabs,array}
\usepackage{calc} 
\usepackage{etoolbox}
\makeatletter
\patchcmd\longtable{\par}{\if@noskipsec\mbox{}\fi\par}{}{}
\makeatother
\IfFileExists{footnotehyper.sty}{\usepackage{footnotehyper}}{\usepackage{footnote}}
\makesavenoteenv{longtable}
\usepackage{graphicx}
\makeatletter
\def\maxwidth{\ifdim\Gin@nat@width>\linewidth\linewidth\else\Gin@nat@width\fi}
\def\maxheight{\ifdim\Gin@nat@height>\textheight\textheight\else\Gin@nat@height\fi}
\makeatother
\setkeys{Gin}{width=\maxwidth,height=\maxheight,keepaspectratio}
\makeatletter
\def\fps@figure{htbp}
\makeatother
\NewDocumentCommand\citeproctext{}{}

\makeatletter
 \let\@cite@ofmt\@firstofone
 \def\@biblabel#1{}
 \def\@cite#1#2{{#1\if@tempswa , #2\fi}}
\makeatother
\newlength{\cslhangindent}
\newlength{\csllabelwidth}
\newenvironment{CSLReferences}[2] 
 {\begin{list}{}{%
  \setlength{\itemindent}{0pt}
  \setlength{\leftmargin}{0pt}
  \setlength{\parsep}{0pt}
  \ifodd #1
   \setlength{\leftmargin}{\cslhangindent}
   \setlength{\itemindent}{-1\cslhangindent}
  \fi
  \setlength{\itemsep}{#2\baselineskip}}}
 {\end{list}}
\usepackage{calc}

\usepackage{url}
\usepackage{glossaries}
\makeglossaries
\usepackage{tikz}
\usetikzlibrary{shapes.geometric, arrows.meta, positioning}
\usetikzlibrary{calc}
\newacronym{ai}{AI}{Artificial Intelligence}
\newacronym{bo}{BO}{Bayesian Optimization}
\newacronym{cart}{CART}{Classification And Regression Tree}
\newacronym{ccd}{CCD}{Central Composite Design}
\newacronym{cnn}{CNN}{Convolutional Neural Network}
\newacronym{cpm}{CPM}{Compressor Performance Map}
\newacronym{cv}{CV}{Cross Validation}
\newacronym{cvfdt}{CVFDT}{Concept-adapting Very Fast Decision Tree}
\newacronym{dace}{DACE}{Design and Analysis of Computer Experiments}
\newacronym{ddm}{DDM}{Drift Detection Method}
\newacronym{dl}{DL}{Deep Learning}
\newacronym{doe}{DOE}{Design of Experiments}
\newacronym{efdt}{EFDT}{Extremely Fast Decision Tree}
\newacronym{gbrt}{gbrt}{Gradient Boosting Regression Tree}
\newacronym{gcd}{GCD}{Greatest Common Divisor}
\newacronym{gra}{GRA}{Global Recurring Abrupt}
\newacronym{hat}{HAT}{Hoeffding Adaptive Tree}
\newacronym{hatc}{HATC}{Hoeffding Adaptive Tree Classifier}
\newacronym{hatr}{HATR}{Hoeffding Adaptive Tree Regressor}
\newacronym{hcf}{HCF}{High-Cycle Fatigue}
\newacronym{hpo}{HPO}{Hyperparameter Optimization}
\newacronym{hpt}{HPT}{Hyperparameter Tuning}
\newacronym{ht}{HT}{Hoeffding Tree}
\newacronym{htc}{HTC}{Hoeffding Tree Classifier}
\newacronym{htr}{HTR}{Hoeffding Tree Regressor}
\newacronym{ig}{IG}{Integrated Gradient}
\newacronym{ki}{KI}{Künstliche Intelligenz}
\newacronym{kpi}{KPI}{Key Performance Indicator}
\newacronym{lasso}{Lasso}{Least Absolute Shrinkage and Selection operator}
\newacronym{mae}{MAE}{Mean Absolute Error}
\newacronym{moo}{MOO}{Multi-Objective Optimization}
\newacronym{ml}{ML}{Machine Learning}
\newacronym{mlp}{MLP}{Multilayer Perceptron}
\newacronym{moa}{MOA}{Massive Online Analysis}
\newacronym{mohpo}{MOHPO}{Multi Objective Hyperparameter Optimization}
\newacronym{mse}{MSE}{Mean Squared Error}
\newacronym{nas}{NAS}{Neural Architecture Search}
\newacronym{nn}{NN}{Neural Network}

\newacronym{ocba}{OCBA}{Optimal Computational Budget Allocation}
\newacronym{pa}{PA}{Passive-Aggressive}
\newacronym{pca}{PCA}{Principal Component Analysis}
\newacronym{rf}{RF}{Random Forest}
\newacronym{rsm}{RSM}{Response Surface Methodology}
\newacronym{river}{river}{River: Online machine learning in Python}
\newacronym{rmoa}{RMOA}{Massive Online Analysis in R}
\newacronym{rocauc}{ROC AUC}{AUC (Area Under The Curve) ROC (Receiver Operating Characteristics)}
\newacronym{sea}{SEA}{SEA synthetic dataset}
\newacronym{shap}{SHAP}{SHapley Additive exPlanations}
\newacronym{sklearn}{sklearn}{scikit-learn: Machine Learning in Python}
\newacronym{smbo}{SMBO}{Surrogate Model Based Optimization}
\newacronym{smote}{SMOTE}{Synthetic Minority Oversampling Technique}
\newacronym{spo}{SPO}{Sequential Parameter Optimization}
\newacronym{spot}{SPOT}{Sequential Parameter Optimization Toolbox}
\newacronym{spotpython}{spotPython}{Sequential Parameter Optimization Toolbox for Python}
\newacronym{spotriver}{spotRiver}{Sequential Parameter Optimization Toolbox for River}
\newacronym{sgd}{SGD}{Stochastic Gradient Descent}
\newacronym{svm}{SVM}{Support Vector Machine}
\newacronym{vfdt}{VFDT}{Very Fast Decision Tree}
\newacronym{xai}{XAI}{Explainable Artificial Intelligence}
\newacronym{xgb}{XGBoost}{eXtreme Gradient Boosting}
\newacronym{blue}{BLUE}{ResponsiBle models, Legal issues, trUst in predictions, Ethical issues}
\newacronym{red}{RED}{Research on data, Explore models, Debug models}
\usepackage{arxiv}
\usepackage{orcidlink}
\usepackage{amsmath}
\usepackage[T1]{fontenc}
\makeatletter
\@ifpackageloaded{caption}{}{\usepackage{caption}}
\AtBeginDocument{%
\ifdefined\contentsname
  \renewcommand*\contentsname{Table of contents}
\else
  \newcommand\contentsname{Table of contents}
\fi
\ifdefined\listfigurename
  \renewcommand*\listfigurename{List of Figures}
\else
  \newcommand\listfigurename{List of Figures}
\fi
\ifdefined\listtablename
  \renewcommand*\listtablename{List of Tables}
\else
  \newcommand\listtablename{List of Tables}
\fi
\ifdefined\figurename
  \renewcommand*\figurename{Figure}
\else
  \newcommand\figurename{Figure}
\fi
\ifdefined\tablename
  \renewcommand*\tablename{Table}
\else
  \newcommand\tablename{Table}
\fi
}
\@ifpackageloaded{float}{}{\usepackage{float}}
\floatstyle{ruled}
\@ifundefined{c@chapter}{\newfloat{codelisting}{h}{lop}}{\newfloat{codelisting}{h}{lop}[chapter]}
\floatname{codelisting}{Listing}

\makeatother
\makeatletter
\@ifpackageloaded{caption}{}{\usepackage{caption}}
\@ifpackageloaded{subcaption}{}{\usepackage{subcaption}}
\makeatother

\ifLuaTeX
  \usepackage{selnolig}  
\fi
\usepackage{bookmark}

\IfFileExists{xurl.sty}{\usepackage{xurl}}{} 
\hypersetup{
  pdftitle={Tuning for Trustworthiness},
  pdfauthor={Alexander Hinterleitner; Thomas Bartz-Beielstein},
  pdfkeywords={XAI, hyperparameter tuning, multi-objective
optimization, desirability function, multi-objective
optimization, surrogate modeling, hyperparameter tuning},
  colorlinks=true,
  linkcolor={blue},
  filecolor={Maroon},
  citecolor={Blue},
  urlcolor={Blue},
  pdfcreator={LaTeX via pandoc}}

\newcommand{\runninghead}{A Preprint }
\title{Tuning for Trustworthiness}
\usepackage{etoolbox}
\makeatletter
\providecommand{\subtitle}[1]{
  \apptocmd{\@title}{\par {\large #1 \par}}{}{}
}
\makeatother
\subtitle{Balancing Performance and Explanation Consistency in Neural
Network Optimization}
\def\asep{\\\\\\ } 
\author{\textbf{Alexander
Hinterleitner}~\orcidlink{0009-0002-7615-6952}\\\\Institute IDE+A, TH
Köln\\Gummersbach,\ 51643\\\href{mailto:alexander.hinterleitner@th-koeln.de}{alexander.hinterleitner@th-koeln.de}\asep\textbf{Thomas
Bartz-Beielstein}~\orcidlink{0000-0002-5938-5158}\\\\Institute IDE+A, TH
Köln\\Gummersbach,\ 51643\\\href{mailto:bartzbeielstein@gmail.com}{bartzbeielstein@gmail.com}}
\date{}
\begin{document}
\maketitle
\begin{abstract}
Despite the growing interest in Explainable Artificial Intelligence
(XAI), explainability is rarely considered during hyperparameter tuning
or neural architecture optimization, where the focus remains primarily
on minimizing predictive loss. In this work, we introduce the novel
concept of XAI consistency, defined as the agreement among different
feature attribution methods, and propose new metrics to quantify it. For
the first time, we integrate XAI consistency directly into the
hyperparameter tuning objective, creating a multi-objective optimization
framework that balances predictive performance with explanation
robustness. Implemented within the Sequential Parameter Optimization
Toolbox (SPOT), our approach uses both weighted aggregation and
desirability-based strategies to guide model selection. Through our
proposed framework and supporting tools, we explore the impact of
incorporating XAI consistency into the optimization process. This
enables us to characterize distinct regions in the architecture
configuration space: one region with poor performance and comparatively
low interpretability, another with strong predictive performance but
weak interpretability due to low \gls{xai} consistency, and a trade-off
region that balances both objectives by offering high interpretability
alongside competitive performance. Beyond introducing this novel
approach, our research provides a foundation for future investigations
into whether models from the trade-off zone---balancing performance loss
and XAI consistency---exhibit greater robustness by avoiding overfitting
to training performance, thereby leading to more reliable predictions on
out-of-distribution data.
\end{abstract}
{\bfseries \emph Keywords}
\def\sep{\textbullet\ }
XAI \sep hyperparameter tuning \sep multi-objective
optimization \sep desirability function \sep multi-objective
optimization \sep surrogate modeling \sep 
hyperparameter tuning

\section{Introduction}\label{introduction}

In recent years, the adoption of \gls{nn} in high-stakes applications
such as healthcare, finance, and autonomous systems has led to a growing
demand for not only high-performing models but also interpretable and
transparent decision-making processes (Carmichael 2024; Zhang et al.
2021; Rudin 2019). While \gls{hpt} and \gls{nas} have primarily focused
on maximizing predictive performance, they often neglect the
explainability of the resulting models. This gap becomes particularly
concerning in domains where trust, accountability, and regulatory
compliance are critical.

\gls{xai} techniques, especially feature attribution methods, have
emerged as a way to interpret the internal decision logic of complex
models. However, these methods are typically applied post hoc and
independently of the optimization process. Evaluating XAI methods
remains a significant challenge, as ground truth for explanations is
often unavailable, making it difficult to assess their predictive
quality and reliability. While there has been increasing effort toward
evaluating feature attribution methods, there is still a lack of
standardized evaluation metrics (Seth and Sankarapu 2025; Longo et al.
2024; Hedström et al. 2023).

In this work, we propose a novel framework that integrates the
consistency of \gls{xai} methods directly into the objective function of
hyperparameter tuning. By formulating the optimization as a
multi-objective problem, we aim to simultaneously improve both model
performance and the explainability of the model. To achieve this, we
define \gls{xai} consistency as the agreement among different feature
attribution scores, reflecting the stability and reliability of
explanations across methods. To quantify this consistency, we introduce
three metrics: the maximum absolute difference, the variance, and the
mean Spearman rank correlation of the feature attribution values.

Our approach is implemented in the \gls{spot}, a flexible and
model-based optimization framework that supports efficient
hyperparameter tuning through surrogate modeling and sequential design
(Bartz-Beielstein 2023). By leveraging \gls{spot}`s capabilities, we are
able to efficiently explore the hyperparameter space while incorporating
\gls{xai} consistency as an explicit optimization objective.

Through different experiments, we investigate the differences in
architectures that emerge when optimization is guided purely by
performance versus when explanation consistency is also considered. Our
results highlight key trade-offs and provide insights into how
\gls{xai}-aware optimization can lead to models that are not only
accurate but also more trustworthy and robust in their interpretability.

This paper is structured as follows: Section~\ref{sec-back} provides
background knowledge on key concepts such as \gls{xai} methods and
techniques for \gls{moo}. Next, Section~\ref{sec-related} reviews
current research in the field of \gls{mohpo} and the application of
\gls{xai} methods in \gls{hpt}. Following that, the methods developed
for this work are introduced in Section~\ref{sec-methods}.
Section~\ref{sec-exp} details the experimental setup, including the
dataset used and the tuning configuration. Subsequently,
Section~\ref{sec-res} presents the results of our study. Finally, we
conclude our work in Section~\ref{sec-sum}.

\section{Background}\label{sec-back}

\subsection{Feature Attribution}\label{feature-attribution}

Feature attribution methods aim to explain machine learning models,
particularly \glspl{nn}, by quantifying the contribution of each input
feature to the model's predictions. These methods differ in their scope
(local vs.~global), timing (post hoc vs.~ante hoc), and level of model
dependency (model-agnostic vs.~model-specific). Understanding these
distinctions is crucial for selecting the right method for a given task.

Local methods explain individual predictions by assigning importance
scores to features for a specific input. Examples include SHAP (Lundberg
and Lee 2017), LIME (Ribeiro, Singh, and Guestrin 2016), and Integrated
Gradients (Sundararajan, Taly, and Yan 2017a). These are particularly
useful for debugging or justifying specific decisions. In contrast,
global methods evaluate overall feature importance across the entire
dataset, such as permutation importance (Fisher, Rudin, and Dominici
2019), providing a broader view of model behavior.

Post hoc methods are applied after the model has been trained, aiming to
explain its decisions without altering the model itself. Most
attribution techniques (including SHAP, LIME, and DeepLIFT (Shrikumar,
Greenside, and Kundaje 2017)) fall into this category. Ante hoc methods,
on the other hand, involve building models that are inherently
interpretable, such as decision trees or generalized additive models.
While ante hoc models are transparent by design, they often lack the
expressive power of deep \glspl{nn}.

Model-agnostic methods (e.g., SHAP, LIME, permutation importance) work
across any type of model by analyzing inputs and outputs without relying
on internal model structures. In contrast, model-specific methods
exploit the internal mechanics of certain architectures, for instance,
DeepLIFT, LRP (Bach et al. 2015), and Integrated Gradients for
\glspl{nn}. These methods tend to be more efficient and accurate when
tailored to a specific model type.

This paper concentrates on local, post hoc attribution methods,
targeting both model-agnostic approaches (Shapley values) and
\gls{nn}-specific techniques (Integrated Gradients, DeepLIFT). These
methods provide detailed insights into how features drive individual
predictions, making them well-suited for interpreting complex models in
sensitive applications.

\subsection{Multi-Objective Optimization with
Desirability}\label{sec-des}

\gls{moo} deals with problems that involve multiple conflicting
objectives, where improving one objective can lead to the degradation of
others. In such cases, it is not possible to find a single solution that
optimizes all objectives simultaneously. To address this, a common
approach is to assign weights to each objective, resulting in weighted
objective functions. These weights reflect the relative importance of
each objective in the context of the optimization problem. The solution
space is then explored to identify a set of Pareto optimal solutions. A
Pareto optimal solution is one in which no objective can be improved
without sacrificing another, and the set of these solutions forms the
Pareto front. The Pareto front represents a trade-off between the
different objectives, offering decision-makers a range of possible
solutions to choose from based on their specific preferences (Derringer
and Suich 1980). In many \gls{moo} problems, determining appropriate
weights for each objective can be challenging and labor-intensive,
particularly when the objectives differ in importance and are measured
on different scales or units. In such cases, the desirability function
approach offers significant advantages. This method not only
standardizes objectives to a common scale, facilitating their
aggregation, but also enables the direct incorporation of
problem-specific preferences and acceptability criteria into the
optimization process. As a result, the desirability approach provides a
more intuitive and flexible framework for handling diverse and
conflicting objectives.

The desirability function, first introduced by Harrington et al. (1965),
maps multiple response variables onto a common 0--1 scale, enabling
simultaneous optimization. In this work, we use the implementation
provided by the \texttt{spotdesirability} package (Bartz-Beielstein
2025a), which adopts the simple discontinuous formulations of Derringer
and Suich (1980). For each response \(f_{r}(x) (r = 1,...,R)\), an
individual desirability \(d_{r}  [0,1]\) dependant to one of the
optimization goal is defined. The different desirability functions can
be combined to calculate one value for the objective function. In our
work we use the geometric mean for this purpose. It can be described by
the following formula:

\[
D = \left( \prod_{r=1}^{R} d_r \right)^{\frac{1}{R}}
\]

Where \(R\) is the number of the desirability functions \(d_r\). The
optimization goals for the different desirability functions \(d_r\) can
be characterized as maximization, minimization and target optimization
problems.

\subsubsection{Maximization}\label{maximization}

If the goal of the optimization is to maximize the objective function,
the desirability problem can be described by the following formula: \[
d_r^{\text{max}} \;=\;
\begin{cases}
0 & \text{if } f_r(x) < A \\[6pt]
\left(\dfrac{f_r(x)-A}{\,B-A\,}\right)^{s} & \text{if } A \le f_r(x) \le B \\[10pt]
1 & \text{if } f_r(x) > B ,
\end{cases}
\]

In this formula, the user must define the following three values:

\begin{itemize}
\tightlist
\item
  \(A\) (Lower Bound): This variable represents the minimum acceptable
  value of \(f_{r}(x)\). If the value of \(f_{r}(x)\) is lower than
  \(A\), the desirability is 0, indicating an unacceptable response.
\item
  \(B\) (Upper Bound): This variable represents the target value where
  desirability reaches the maximum. Responses of \(f_{r}(x)\) above
  \(B\) are considered as equally ideal with a desirability value of 1.
\item
  s (Steepness): This variable controls the desirability steepness
  between \(A\) and \(B\). Values of 1 result in a linear relationship
  between \(f_{r}(x)\) and the desirability. If \(s <1\) the curve
  between \(A\) and \(B\) is concave, making it easier to achieve higher
  desirability even if \(f_{r}(x)\) is not close to \(B\). If \(s >1\)
  the curve between \(A\) and \(B\) is convex. This requires
  \(f_{r}(x)\) to be close to \(B\) to achieve high desirability.
\end{itemize}

\subsubsection{Minimization}\label{minimization}

The goal in this case is minimization and can be described by the
following formula:

\[
d_r^{\text{min}} =
\begin{cases}
1 & \text{if } f_r(x) < A \\[6pt]
\left(\dfrac{B - f_r(x)}{B - A}\right)^{s} & \text{if } A \le f_r(x) \le B \\[10pt]
0 & \text{if } f_r(x) > B ,
\end{cases}
\]

In this formula, the user must define the following three values:

\begin{itemize}
\tightlist
\item
  \(A\) (Lower Bound): This variable represents the target value of
  \(f_r(x)\). If the value of \(f_r(x)\) is lower than \(A\), the
  desirability is 1, indicating a fully desirable response.
\item
  \(B\) (Upper Bound): This variable represents the highest acceptable
  value of \(f_{r}(x)\) where desirability reaches its minimum.
  Responses of above \(B\) are considered unacceptable, with a
  desirability value of 0.
\item
  \(s\) (Steepness): This controls the desirability steepness between
  \(A\) and \(B\). Values of \(s = 1\) result in a linear relationship
  between \(f_r(x)\) and desirability. For \(s < 1\) the curve between
  \(A\) and \(B\), it easier to achieve higher desirability even if
  \(f_{r}(x)\) is not close to \(B\).
\end{itemize}

\subsubsection{Target Optimization}\label{target-optimization}

In situation where the optimization target is to achieve a specific
value, the following function is used:

\[
d_r^{\text{target}} =
\begin{cases}
\left( \dfrac{f_r(x) - A}{t_0 - A} \right)^{s_1} & \text{if } A \leq f_r(x) \leq t_0 \\[12pt]
\left( \dfrac{f_r(x) - B}{t_0 - B} \right)^{s_2} & \text{if } t_0 \leq f_r(x) \leq B \\[12pt]
0 & \text{otherwise}
\end{cases}
\]

Where:

\begin{itemize}
\tightlist
\item
  \(A\) (Lower Bound): This variable defines the minimum accepted value
  for the response. If the response is below \(A\), the desirability is
  0, indicating an unacceptable outcome.
\item
  \(B\) (Upper Bound): This variable defines the maximum accepted value
  for the response. If the response exceeds \(B\), the desirability is
  0, indicating an unacceptable outcome.
\item
  \(t_0\) (Target Value): This variable represents the most desirable
  value for the response, where the desirability reaches 1.
\item
  \(s_1\) (Steepness for \(A \leq f_r(x) \leq t_0\)): This parameter
  controls the steepness of the desirability curve for values between
  \(t_0\) and \(A\).
\item
  \(s_2\) (Steepness for \(t_0 \leq f_r(x) \leq B\)): This parameter
  controls the steepness of the desirability curve for values between
  \(t_0\) and \(B\).
\end{itemize}

\section{Related Work}\label{sec-related}

The evolution of \gls{hpo} from single-objective to multi-objective
frameworks reflects the increasing complexity and practical demands of
machine learning systems. In real-world deployments, predictive accuracy
alone is often insufficient. Further objectives such as computational
efficiency, calculation time, and environmental impact are gaining
prominence. Hennig and Lindauer (n.d.) exemplify this transition by
combining multi-fidelity \gls{hpo} with \gls{moo} to balance accuracy
and energy efficiency in Deep Shift Neural Networks. By leveraging shift
operations to reduce computational costs, these networks enable
Pareto-optimal trade-offs between performance loss and carbon emissions,
demonstrating that simultaneous improvements in both dimensions are
feasible. Their work reflects a broader trend in \gls{mohpo}, where
diverse criteria increasingly shape the optimization landscape.

Karl et al. (2023) provide a detailed taxonomy of \gls{mohpo}
approaches, highlighting evolutionary algorithms and Bayesian
optimization as the dominamt strategies. Evolutionary algorithms are
particularly effective in exploring diverse regions of the search space,
making them suitable for problems with multiple competing objectives.
Bayesian optimization methods, utilize surrogate models to efficiently
navigate complex, high-dimensional spaces. These techniques are
especially valuable in hardware-aware neural architecture design, where
performance must be balanced against constraints such as energy
consumption and latency. For example, Parsa et al. (2020) introduce
Hierarchical Pseudo Agent-based Bayesian Optimization, a framework that
jointly considers model accuracy, energy usage, and hardware
limitations, thus bridging the gap between software performance and
hardware feasibility.

A persistent challenge in \gls{mohpo} is the formulation of meaningful
quality indicators for comparing Pareto fronts. Giovanelli et al. (2024)
address this through an interactive preference learning approach that
replaces static performance metrics with user-guided comparisons. By
learning a latent utility function from pairwise preferences over
Pareto-optimal solutions, their method captures human-centric priorities
dynamically.

Despite these advances, current \gls{mohpo} frameworks largely overlook
\gls{xai} consistency as an optimization objective, but there are first
research progressing in this field. Chakraborty, Seifert, and Wirth
(2024) leverages XAI not to explain the prediction model itself, but
rather to interpret the surrogate models used within the Bayesian
optimization process. By providing explanations for the surrogate's
decision-making, the authors aim to make the hyperparameter search more
transparent and efficient. In this context, \gls{xai} serves as a tool
for understanding and guiding the optimization process, rather than for
interpreting the final prediction model.

Sumita, Nakagawa, and Tsuchiya (2023) focus on enhancing hyperparameter
tuning for deep learning models in time-series forecasting by
incorporating \gls{xai} techniques. The method utilizes explainability
to analyze the influence of different hyperparameters, such as window
size, on both model performance and prediction outcomes. By interpreting
these influences, Xtune aims to accelerate and improve the reliability
of the hyperparameter search.

Chandramouli, Zhu, and Oulasvirta (2023) apply XAI analysis directly to
the prediction classifier, optimizing not only for accuracy but also for
explainability as evaluated by human users. The approach involves an
interactive loop in which users iteratively assess the quality of model
explanations, and this feedback is incorporated as a secondary objective
in multi-objective Bayesian optimization.

Unlike previous work, our approach does not seek to validate the
explanations themselves or require human-in-the-loop feedback. Instead,
we incorporate the explainability of the \gls{nn} as a second objective
during hyperparameter tuning by leveraging XAI consistency metrics. In
this way, we aim to guide the hyperparameter optimization process toward
more transparent and interpretable \glspl{nn}.

\section{Methods}\label{sec-methods}

\subsection{Multi-Objective Hyperparameter Tuning With
Spotpython}\label{multi-objective-hyperparameter-tuning-with-spotpython}

\gls{smbo} has become a key element in tackling complex simulation and
optimization tasks, particularly when dealing with expensive black-box
functions. In the context of \gls{moo}, where several conflicting
objectives must be balanced, the need for efficient, reliable, and
interpretable optimization methods is even more pronounced.

\gls{spot}, available as the open-source Python package
\texttt{spotpython}, addresses these challenges by offering a flexible
and robust framework for single- and multi-objective
optimization\footnote{https://github.com/sequential-parameter-optimization/}.
\gls{spot} supports advanced statistical modeling techniques and
provides a structured environment for handling both the exploration and
exploitation phases critical to \gls{moo} scenarios. This is supported
by the option of using individual weighted aggregation functions as well
as implementations for determining the pareto front or using
desirability approaches (Bartz-Beielstein 2025b).

At its core, \gls{spot} employs surrogate models to approximate the
objective functions, significantly reducing the number of expensive
evaluations needed (Bartz et al. 2023). It supports a variety of
surrogate modeling techniques, including classical regression models,
analysis of variance (ANOVA), and modern machine learning approaches
such as classification and regression trees (CART), random forests, and
Gaussian processes (Kriging) (Krige 1951). These models are well-suited
for capturing the trade-offs among multiple objectives. A key strength
of SPOT lies in its modular design, which allows users to integrate and
combine various meta-modeling strategies. \gls{spot}'s compatibility
with any model from scikit-learn (Pedregosa et al. 2011) further extends
its versatility, enabling users to experiment with cutting-edge
surrogate modeling techniques tailored to specific problem
characteristics.

\begin{figure}[htbp]
\centering
\begin{tikzpicture}[node distance=1.6cm]

\tikzstyle{block} = [rectangle, draw, rounded corners, align=center, minimum width=6.2cm, minimum height=1.2cm]
\tikzstyle{arrow} = [thick,->,>=stealth]

\node (init) [block] {Init:\\ Build initial design $X$};
\node (evaluate) [block, below of=init] {Evaluate initial design on real objectives:\\ $\mathbf{Y} = \mathbf{f}(X) = [f_1(X), f_2(X), \dots, f_k(X)]$};
\node (surrogate) [block, below of=evaluate] {Build surrogate model:\\ $\mathbf{S} = S(X, \mathbf{Y})$};
\node (optimize) [block, below of=surrogate] {Optimize surrogate :\\ $X_0 = \text{optimize}(\mathbf{S})$};
\node (evaluate0) [block, below of=optimize] {Evaluate on real objectives:\\ $\mathbf{Y}_0 = \mathbf{f}(X_0)$};
\node (infill) [block, below of=evaluate0] {Infill new points:\\ $X = X \cup X_0$, $\mathbf{Y} = \mathbf{Y} \cup \mathbf{Y}_0$};
\node (loop) [block, below of=infill] {Check if stopping criterion is met};

\draw [arrow] (init) -- (evaluate);
\draw [arrow] (evaluate) -- (surrogate);
\draw [arrow] (surrogate) -- (optimize);
\draw [arrow] (optimize) -- (evaluate0);
\draw [arrow] (evaluate0) -- (infill);
\draw [arrow] (infill) -- (loop);
\draw [arrow] 
  (loop.south) -- ++(0,-0.8)
  -- ($ (loop.south)+(-4.5,-0.8) $)
  -- ($ (surrogate.west)+(-1.4,0) $)
  -- (surrogate.west);
\node[left] at ($(loop.south)!0.5!($ (loop.south)+(-3.5,-2.1) $)$) {If not};

\end{tikzpicture}
\caption{Workflow of the surrogate model-based optimization algorithm in \gls{spot}. The process alternates between building surrogate models and optimizing them, with new candidate solutions evaluated on the true objectives and added to the dataset until a stopping criterion is met.}
\label{fig:surrogate-workflow}
\end{figure}
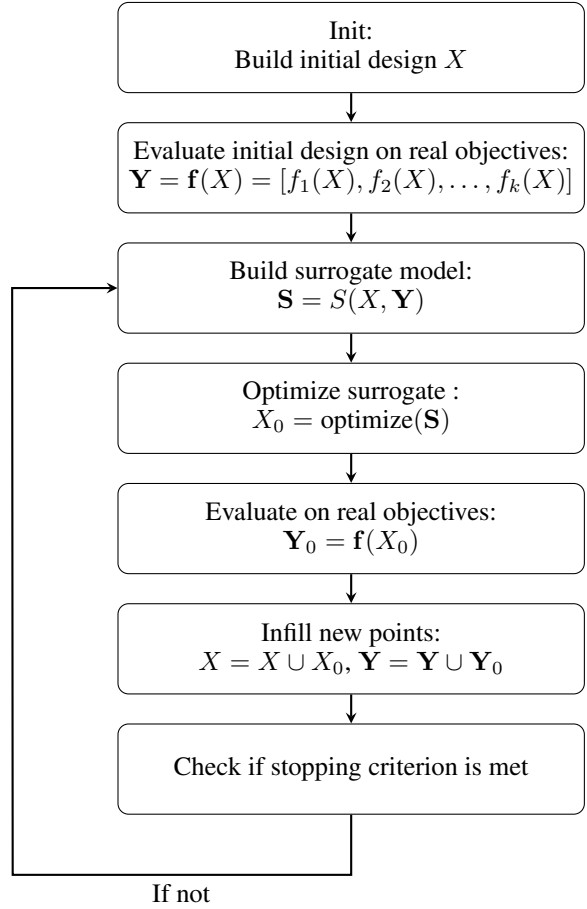

\gls{spot} is designed to handle both discrete and continuous decision
variables, making it applicable to a wide range of optimization
problems(Zaefferer, Stork, and Bartz-Beielstein 2014; Bartz-Beielstein
and Zaefferer 2017). In addition, it incorporates exploratory fitness
landscape analysis and sensitivity analysis tools, which help users
understand the behavior and interactions of objectives and parameters.
One of \gls{spot}'s notable applications is hyperparameter tuning,
including for \gls{moo} problems, where objectives such as accuracy,
training time, or, in our case, consistency of explanations must be
optimized simultaneously. The surrogate-based optimization steps for
\gls{moo} are illustrated in Figure \ref{fig:surrogate-workflow}.

\subsection{XAI Consistency}\label{xai-consistency}

A fundamental aspect of this work is addressing the challenge of
evaluating the consistency of \gls{xai} methods for \gls{nn}
interpretability. Local explanation techniques assign feature
attributions to individual predictions, but assessing their reliability
is difficult in the absence of ground truth. Instead of evaluating the
quality of a single explanation, we focus on the consistency among
different feature attribution methods. The underlying assumption is that
when different \gls{xai} methods consistently highlight the same
important features, this suggests that the model has learned meaningful
patterns, rather than relying on spurious correlations or dataset
artifacts (Kozielski, Sikora, and Wawrowski 2025; Kazmierczak et al.
2024). In this context, our goal is to guide the hyperparameter tuning
search process toward model architectures that exhibit a higher degree
of explainability.

To systematically study \gls{xai} consistency, we compare three widely
used and theoretically distinct local explanation methods: DeepLIFT,
Integrated Gradients, and KernelSHAP. These methods were chosen
specifically for their diverse conceptual foundations, enabling a broad
and representative analysis of explanation behavior.

\textbf{Deep Learning Important FeaTures (DeepLIFT)} is a model-specific
feature attribution method for \glspl{nn} that attributes the prediction
of a model by comparing the activation of neurons to a reference
input(Shrikumar, Greenside, and Kundaje 2017). Unlike standard
gradients, it propagates the difference in activations, which helps
avoid problems like vanishing gradients. It computes the contribution of
each input feature vector \(x\) as the difference in output relative to
a reference \(x'\), while preserving a summation-to-delta property.

\textbf{Integrated Gradients} is a path-based attribution method
designed for differentiable models (Sundararajan, Taly, and Yan 2017b).
It attributes feature importance by integrating the gradients of the
model's output along a straight-line path from a baseline input \(x'\)
to the actual input \(x\). The attribution for input feature \(x_i\) is
computed as:

\[ \text{IG}_i(x) = (x_i - x'_i) \times \int_{\alpha=0}^{1} \frac{\partial f(x' + \alpha(x - x'))}{\partial x_i} \, d\alpha
\]

Here, \(\alpha \in [0, 1]\) is a scaling factor that determines the
position along the straight-line path between the baseline \(x'\) and
the input \(x\). Specifically:

\begin{itemize}
\tightlist
\item
  \(\alpha = 0\) corresponds to the baseline \(x'\),
\item
  \(\alpha = 1\) corresponds to the actual input \(x\),
\item
  Intermediate values \(0 < \alpha < 1\) represent partial progress
  along this path.
\end{itemize}

This method satisfies the following key axioms:

\begin{itemize}
\tightlist
\item
  Sensitivity: Attributions reflect actual differences in model outputs.
\item
  Implementation Invariance: Functionally equivalent models yield the
  same attributions.
\end{itemize}

\textbf{KernelSHAP} is a model-agnostic, perturbation-based attribution
method grounded in cooperative game theory (Lundberg and Lee 2017). It
approximates Shapley values, which represent the average marginal
contribution of each input feature across all possible feature
subsets(Shapley et al. 1953). These values are derived from the concept
of fair payout distribution in coalitional games, applied here to
quantify feature importance. The Shapley value for feature \(x_i\) is
defined as:

\[
\phi_{x_i} = \sum_{S \subseteq N \setminus \{x_i\}} \frac{|S|! (|N| - |S| - 1)!}{|N|!} \left[ f(S \cup \{x_i\}) - f(S) \right]
\]

where:

\begin{itemize}
\tightlist
\item
  \(N\) is the set of all input features,
\item
  \(S\) is a subset of features excluding \(x_i\),
\item
  \(f(S)\) is the model prediction when only the features in \(S\) are
  known.
\end{itemize}

Computing exact Shapley values is computationally infeasible for models
with many features. In such cases, KernelSHAP offers an efficient
approximation by applying a locally weighted linear regression over a
set of randomly sampled feature subsets (i.e., perturbed versions of the
input). The weights used in the regression reflect the importance of
each subset in approximating the Shapley kernel, ensuring that the
resulting attributions remain faithful to the theoretical properties of
Shapley values.

\subsubsection{Consistency Metrices}\label{sec-cons}

We aggregate the mean of all local explanations for each feature to
derive a global feature importance profile for each \gls{xai} method.
This aggregated view serves as the foundation for our consistency
analysis across methods. Since the evaluation of feature attribution
techniques remains a challenging research problem, primarily due to the
lack of an objective ground truth for explanations, we propose a set of
metrics to assess the consistency between different \gls{xai} methods.
These metrics aim to provide an empirical basis for comparing
attribution outputs and identifying stable patterns of feature
importance.

To measure the consistency between different \gls{xai} methods, we
implemented three metrics in \texttt{spotpython}. The first metric is
the sum of the maximum absolute differences between the explanations.
This metric is related to the largest disagreement among the \gls{xai}
methods. The value of the metric can theoretically be dominated by a
single large difference for one feature. The metric can be described by
the following formula:

\[
Cons_{max\_diff} = \sum_{j=1}^m \max_{i,k \in \{1, \ldots, n\}} \left| E[i, j] - E[k, j] \right|
\]

where:

\begin{itemize}
\tightlist
\item
  \(E[i, j]\) is the value of the \(j\)-th feature in the \(i\)-th
  explanation,
\item
  \(n\) is the number of feature attribution methods,
\item
  \(m\) is the number of features.
\end{itemize}

The second metric is the sum of the variance for each feature across all
explanations from the different feature attribution methods. This metric
captures the overall disagreement among all \gls{xai} methods. It is
sensitive to outliers and does not account for the order of the feature
importances. The sum of the \gls{xai} variances can be expressed by the
following formula:

\[
Cons_{var} = \sum_{j=1}^m \text{Var}_j \left( E[1, j], E[2, j], \dots, E[n, j] \right)
\]

The final metric implemented in \texttt{spotpython} is the mean of the
Spearman rank correlation (Spearman 1961) coefficients across all
feature attribution methods. This metric measures the consistency in the
ranking order of feature importance values estimated by the different
\gls{xai} methods. Unlike the other metrics, the Spearman rank
correlation does not consider the magnitude of the explanations, making
it more robust to outliers. Its ability to capture agreement in feature
ranking is especially valuable when the relative importance of features
is more meaningful than their exact scores. The metric is given by the
following formula:

\[
Cons_{spearman} = \frac{2}{n(n-1)} \sum_{1 \leq i < k \leq n} \rho(E_i, E_k)
\]

where:

\begin{itemize}
\tightlist
\item
  \(\rho(\mathbf{E}_i, \mathbf{E}_k)\) is the Spearman rank correlation
  coefficient between the \(i\)-th and \(k\)-th explanation vectors. It
  is defined by:
\end{itemize}

\[
\rho(E_i, E_k) = 1 - \frac{6 \sum_{j=1}^m (r_{i,j} - r_{k,j})^2}{m(m^2 - 1)}
\]

where:

\begin{itemize}
\tightlist
\item
  \(r_{i,j}\) and \(r_{k,j}\) are the ranks of the \(j\)-th feature in
  the \(i\)-th and \(k\)-th explanation vectors, respectively,
\item
  \(m\) is the number of features.
\end{itemize}

\section{Experiments}\label{sec-exp}

In this study, we compare three tuning approaches with respect to their
impact on explanation consistency, performance loss, and the resulting
\gls{nn} architecture. The first approach optimizes the network's
hyperparameters based on an objective function that minimizes
performance loss alone. The second approach extends this by
incorporating explanation consistency, quantified using Spearman
correlation as described in Section~\ref{sec-cons}, into the objective
function equally weighted as the performance loss. The third approach
further develops this idea by employing a desirability function, which
maps both performance loss and explanation consistency (also measured
via Spearman correlation) onto a normalized objective value between 0
and 1, based on predefined desirable ranges, as detailed in
Section~\ref{sec-des}.

\subsection{Data}\label{data}

For demonstration purposes, we use the California Housing dataset from
scikit-learn, which can be loaded using the
\texttt{fetch\_california\_housing} function (Pedregosa et al. 2011).
The dataset comprises eight features that describe various attributes of
houses in California, for example the age of the house, or its
geographic location (longitude and latitude). The target variable
represents the median house value, ranging from 0.15 to 5.0, where the
values are scaled in units of \$100,000. This corresponds to actual
house prices between \$15,000 and \$500,000. Accordingly, the dataset is
used to demonstrate our approach in the context of a regression problem.
The dataset is split into three subsets: 60\% for training, 20\% for
validation, and 20\% for testing. The validation set is used to evaluate
the objective function during the tuning process, while the test set
remains completely untouched to ensure unbiased performance assessment
after tuning. During the tuning process, the data is standardized using
a custom implementation of the standard scaler. The scaler is fitted on
the training data and then applied to transform the validation data
accordingly.

\subsection{Tuning Setup}\label{sec-setup}

For the tuning process, we use a wrapper class for a regression \gls{nn}
implemented in the \texttt{spotpython} framework
(\texttt{NetLightRegression}). This class defines a \gls{mlp} with four
hidden layers. The input layer matches the dimensionality of the feature
space. The number of neurons in the first hidden layer is treated as a
tunable hyperparameter. The second and third hidden layers each contain
half as many neurons as the first, while the fourth layer contains one
quarter. Each hidden layer is followed by a dropout layer to reduce the
risk of overfitting. The output layer is a linear layer with a single
neuron, as the model is designed to predict a single continuous target
variable.

During hyperparameter tuning, we optimize several parameters: the number
of neurons in the first hidden layer (\texttt{l1}), the number of
training epochs, batch size, dropout probability, activation functions
for the layers, the optimizer, and a learning rate multiplier. Instead
of tuning the learning rate directly, \texttt{spotpython} tunes a
learning rate multiplier. This approach is used because the optimal
learning rate scale can vary depending on the chosen optimizer. For
example, the learning rate for the Adagrad optimizer is computed as
\(learning\_rate \times 0.01\), whereas for the Adam optimizer, it is
computed as \(learning\_rate \times 0.001\). The boundaries for the
hyperparameter tuning runs can be seen in Table~\ref{tbl-hyperparams}.

\begin{longtable}[]{@{}
  >{\raggedright\arraybackslash}p{(\columnwidth - 8\tabcolsep) * \real{0.2625}}
  >{\raggedright\arraybackslash}p{(\columnwidth - 8\tabcolsep) * \real{0.1750}}
  >{\raggedright\arraybackslash}p{(\columnwidth - 8\tabcolsep) * \real{0.1875}}
  >{\raggedright\arraybackslash}p{(\columnwidth - 8\tabcolsep) * \real{0.1875}}
  >{\raggedright\arraybackslash}p{(\columnwidth - 8\tabcolsep) * \real{0.1875}}@{}}
\caption{Hyperparameter search space and transformations used for the
tuning runs}\label{tbl-hyperparams}\tabularnewline
\toprule\noalign{}
\begin{minipage}[b]{\linewidth}\raggedright
Hyperparameter
\end{minipage} & \begin{minipage}[b]{\linewidth}\raggedright
Lower Bound
\end{minipage} & \begin{minipage}[b]{\linewidth}\raggedright
Upper Bound
\end{minipage} & \begin{minipage}[b]{\linewidth}\raggedright
Transformation
\end{minipage} & \begin{minipage}[b]{\linewidth}\raggedright
Options
\end{minipage} \\
\midrule\noalign{}
\endfirsthead
\toprule\noalign{}
\begin{minipage}[b]{\linewidth}\raggedright
Hyperparameter
\end{minipage} & \begin{minipage}[b]{\linewidth}\raggedright
Lower Bound
\end{minipage} & \begin{minipage}[b]{\linewidth}\raggedright
Upper Bound
\end{minipage} & \begin{minipage}[b]{\linewidth}\raggedright
Transformation
\end{minipage} & \begin{minipage}[b]{\linewidth}\raggedright
Options
\end{minipage} \\
\midrule\noalign{}
\endhead
\bottomrule\noalign{}
\endlastfoot
\textbf{l1} & 2.0 & 10.0 & \(2^x\) & - \\
\textbf{epochs} & 4.0 & 11.0 & \(2^x\) & - \\
\textbf{batch size} & 4.0 & 10.0 & \(2^x\) & - \\
\textbf{dropout probability} & 0.0 & 0.4 & - & - \\
\textbf{learning rate multiplier} & 0.1 & 5.0 & - & - \\
\textbf{activation function} & - & - & - & \texttt{ReLU,LeakyReLU,} \\
& & & & \texttt{ELU,Swish} \\
\textbf{optimizer} & - & - & - & \texttt{Adam,Adamax,} \\
& & & & \texttt{SGD,\ NAdam,} \\
& & & & \texttt{RAdam,Adagrad,} \\
& & & & \texttt{RMSprop} \\
\end{longtable}

For the general tuning setup across all three experiments, we employ
\gls{smbo} using a Kriging model optimized with Differential Evolution
(Storn and Price 1997). The initial design, which consists of points in
the design space that are evaluated to construct the first surrogate
model, includes 20 points. These points are distributed within the
design space using Latin Hypercube Sampling (Leary, Bhaskar, and Keane
2003). After constructing the initial surrogate model, the tuning
process begins with a maximum budget of 60 function evaluations. To
reduce the impact of potential outliers or random fluctuations in
performance, each design point is evaluated twice, and the mean of the
two results is used for further optimization. For the performance loss
calculation in this demonstration, the \gls{mse} is used, as it
penalizes outliers more strongly than metrics such as \gls{mae} or the
coefficient of determination (\(R^2\)). However, several other
performance metrics are also implemented in \texttt{spotpython},
providing flexibility depending on the specific requirements of the
tuning task.

In the first baseline experiment, only the performance loss is used as
the optimization objective. In the second experiment, we extend the
objective by also incorporating the mean Spearman rank correlation as a
measure of explanation consistency. We chose this metric for
demonstration purposes, as in many real-world applications it is more
relevant to identify the most important features, or their relative
ranking, than to rely on the exact magnitude of the attribution
values.The consistency is calculated across three feature attribution
methods: Integrated Gradients, DeepLift, and KernelSHAP. Since our goal
is solely to demonstrate the consistency analysis of different feature
attribution methods, we set the baseline to a null vector. Both
performance loss and explanation consistency are weighted equally in
this setup. Since the consistency scores range between 0 and 1, and the
target values range from 0.15 to 5.0, this equal weighting provides a
plausible balance, giving both components similar importance.

\begin{figure}

\begin{minipage}{0.50\linewidth}
\includegraphics{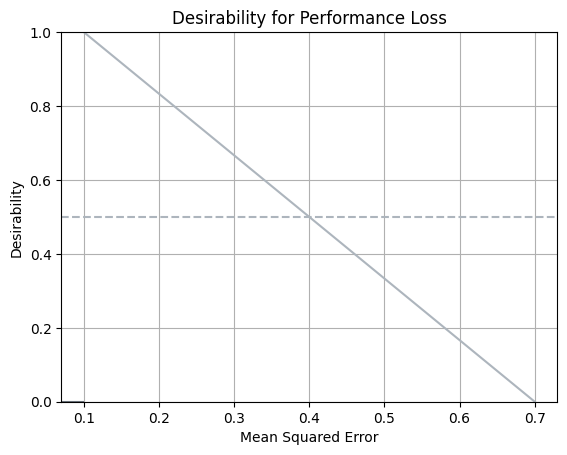}\end{minipage}%
\begin{minipage}{0.50\linewidth}
\includegraphics{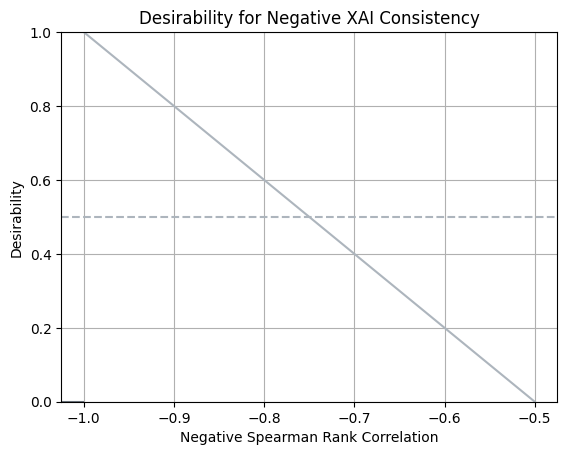}\end{minipage}%

\caption{\label{fig-desire}Desirability functions used for tuning:
(Left) Desirability of the performance loss. Values $\leq$ 0.1 achieve the
highest desirability, while values $\geq$ 0.7 are considered undesirable.
(Right) Desirability of the negative XAI consistency loss. Values are
negated due to the \texttt{spotpython} minimization setup. A value of
--1 achieves the highest desirability, while values $\geq$ --0.5 are
considered undesirable.}

\end{figure}

However, as performance loss decreases (improves) and explanation
consistency increases, the consistency score begins to dominate the
overall objective function. To address this imbalance, we introduce a
third experiment based on a desirability function. In this approach, we
define a desirability range for the performance loss: values below 0.1
are considered highly desirable, while values above 0.7 are treated as
undesirable. For the \gls{xai} consistency, a value of --1 is defined as
the most desirable outcome. The negative sign is used to align with the
\texttt{spotpython} framework, which is designed for minimization
problems. Input values above -0.5, indicating a consistency of 50\% or
less in the ranking of feature importance across attribution methods,
are considered undesirable. Between these two thresholds, desirability
increases linearly, as illustrated in Figure~\ref{fig-desire}. As
discussed in Section~\ref{sec-des}, modifying the \(s\) parameter can
also result in non-linear desirability functions; however, for
demonstration purposes, we use a linear configuration in this study.

Both desirability functions are combined and mapped to a common scale
between 0 and 1. If either component of the objective function falls
into the undesirable range, the overall desirability is set to 0. Since
the \texttt{spotpython} framework is designed for minimization tasks,
the final objective value is computed as \(1 - desirability\) ,
effectively turning the maximization of desirability into a minimization
problem.

\section{Results}\label{sec-res}

\begin{figure}

{\centering \includegraphics[width = 0.9\textwidth]{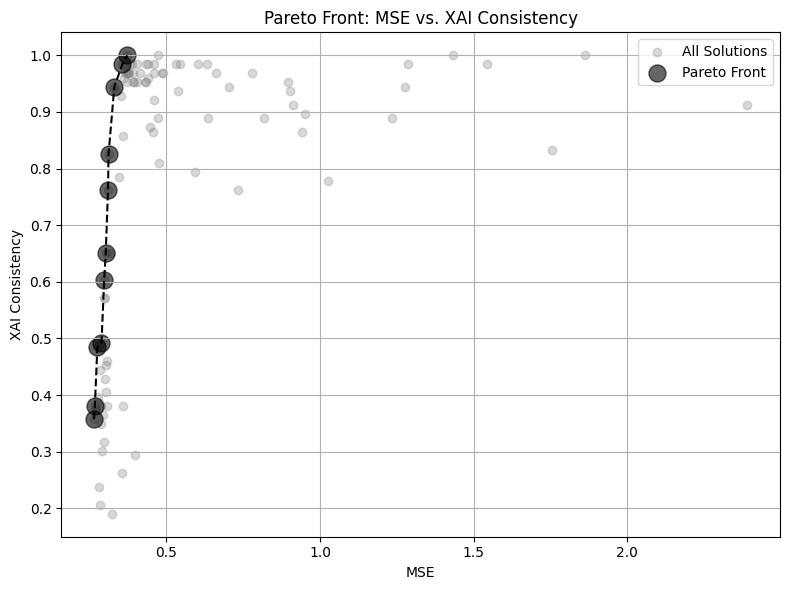}

}

\caption{Pareto plot of a Latin hypercube sampling of 100 points in the
design space. The y-axis represents XAI consistency values, while the
x-axis shows the MSE values for different tuning configurations. The
Pareto front and its corresponding points are marked in black, while the
remaining configurations are shown as grey points. Configurations with
MSE values above 3 are excluded to improve visualization of the relevant
regions, resulting in 94 remaining points in the design space.
\label{fig-pareto}}

\end{figure}

To gain an overview of how the different \gls{nn} configurations behave
in terms of \gls{xai} consistency and performance loss, we generated a
Pareto plot based on the evaluations of 100 designs selected using Latin
hypercube sampling to cover the design space (see Figure
\ref{fig-pareto}). This initial visualization was generated without
using any configurations obtained through \gls{hpt}. The black points
connected by a black dotted line represent the Pareto points and the
Pareto front. The gray dots correspond to all other design points.
Configurations with \gls{mse} values above 3 are excluded to improve the
visualization of relevant regions, resulting in 94 remaining points in
the design space.

One notable observation is the empty area in the bottom-right corner,
indicating that no configurations in the design space result in both
very low \gls{xai} consistency values and high \gls{mse} values. All
\gls{mse} values above 0.5 correspond to configurations that achieve at
least 75\% \gls{xai} consistency.The Pareto front itself appears as a
nearly vertical line, but the configuration with the best MSE (around
0.2) exhibits poor \gls{xai} consistency, with an average Spearman rank
correlation of approximately 35\%. However, it is also shown that
achieving XAI consistency above 95\% does not require major sacrifices
in terms of performance loss.

To compare the three different tuning approaches, namely single
objective tuning based on the \gls{mse}, and two multi-objective
approaches combining \gls{mse} and \gls{xai} consistency (one with equal
weighting and the other using a desirability function), we evaluated
their performance on both the validation data used during the tuning
process and the unseen test data, which was not involved in tuning.

\begin{figure}[H]

{\centering \includegraphics{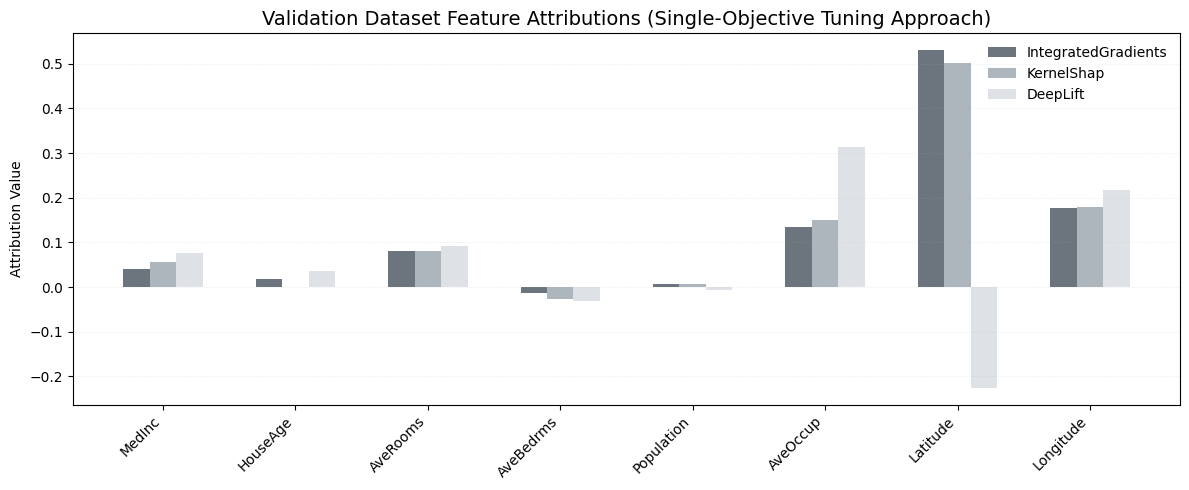}

}

\caption{Feature attribution values computed using different XAI methods
on the validation dataset, after tuning the network architecture based
solely on performance loss (single-objective tuning). \label{fig-fa-so}}

\end{figure}

\begin{figure}[H]

{\centering \includegraphics{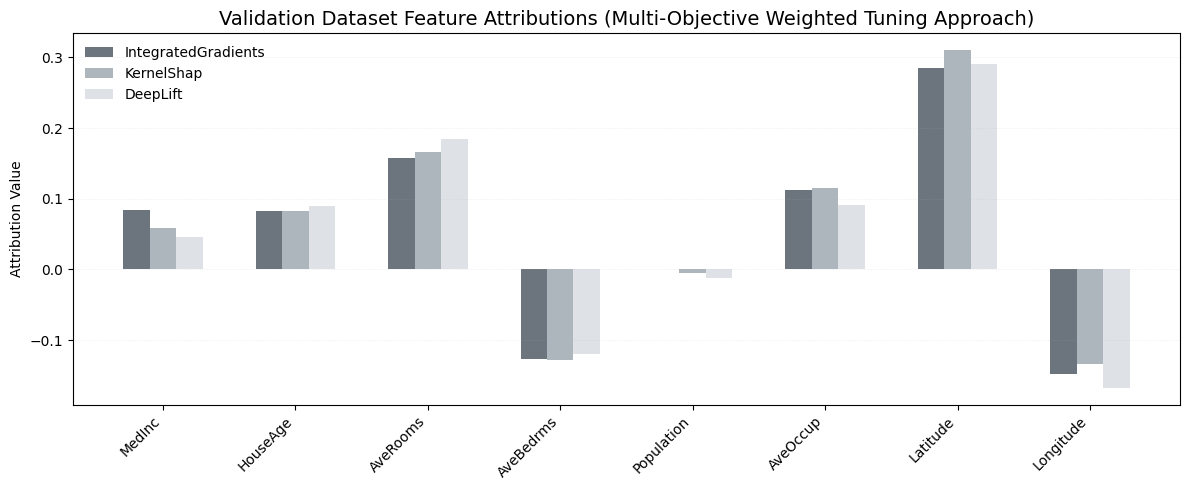}

}

\caption{Feature attribution values computed using different XAI methods
on the validation dataset, after tuning the network architecture with
equal weighting of XAI consistency and performance loss (multi-objective
weighted approach). \label{fig-fa-mow}}

\end{figure}

The results are presented in Table~\ref{tbl-res}. It is evident that the
\gls{xai} consistency---measured by the Spearman rank
correlation---remains stable above 98\% for the two multi-objective
tuning approaches on both the validation and test datasets. In contrast,
the single-objective loss-based tuning approach results in varying
feature importances across different attribution methods, as reflected
by lower Spearman rank correlation values: 52.4\% for the validation
data and 82.6\% for the test data. This behavior is also observable in
the visual comparison of feature attribution values on the validation
data between the single-objective loss-based tuning (see Figure
\ref{fig-fa-so}) and the multi-objective weighted approach (see Figure
\ref{fig-fa-mow}). It can be observed that the feature attribution
methods applied to the model tuned solely on performance loss yield
differing importance scores, particularly for the feature Average
Occupants per Household (AveOccup). For the feature representing the
geographical latitude of the house (Latitude), Integrated Gradients and
KernelSHAP indicate a strong positive impact on house price, whereas
DeepLIFT estimates a strong negative impact. In contrast, the
attribution values generated by different methods for the model tuned
using the equally weighted multi-objective approach show consistent
importance scores.

The \gls{mse} values on the validation data during tuning show similar
results across all three tuning experiments (see Table~\ref{tbl-res}).
Interestingly, the model tuned using the multi-objective desirability
approach achieves slightly better performance on the validation data
compared to the model tuned solely based on performance loss. However,
on the unseen test data, the loss-based tuning approach yields a lower
\gls{mse} than both multi-objective approaches.

\begin{longtable}[]{@{}
  >{\raggedright\arraybackslash}p{(\columnwidth - 6\tabcolsep) * \real{0.4000}}
  >{\raggedright\arraybackslash}p{(\columnwidth - 6\tabcolsep) * \real{0.2000}}
  >{\raggedright\arraybackslash}p{(\columnwidth - 6\tabcolsep) * \real{0.2143}}
  >{\raggedright\arraybackslash}p{(\columnwidth - 6\tabcolsep) * \real{0.1857}}@{}}
\caption{Results for XAI consistency (higher values are better) and MSE
(lower values are better) on validation and test data across the three
tuning experiments: single-objective loss-based, multi-objective
weighted, and multi-objective
desirability-based.}\label{tbl-res}\tabularnewline
\toprule\noalign{}
\begin{minipage}[b]{\linewidth}\raggedright
Metric
\end{minipage} & \begin{minipage}[b]{\linewidth}\raggedright
Loss-Based
\end{minipage} & \begin{minipage}[b]{\linewidth}\raggedright
Weighted
\end{minipage} & \begin{minipage}[b]{\linewidth}\raggedright
Desirability
\end{minipage} \\
\midrule\noalign{}
\endfirsthead
\toprule\noalign{}
\begin{minipage}[b]{\linewidth}\raggedright
Metric
\end{minipage} & \begin{minipage}[b]{\linewidth}\raggedright
Loss-Based
\end{minipage} & \begin{minipage}[b]{\linewidth}\raggedright
Weighted
\end{minipage} & \begin{minipage}[b]{\linewidth}\raggedright
Desirability
\end{minipage} \\
\midrule\noalign{}
\endhead
\bottomrule\noalign{}
\endlastfoot
\textbf{XAI Consistency (Validation Data)} & 0.524 & 0.984 & 0.984 \\
\textbf{XAI Consistency (Test Data)} & 0.826 & 1.0 & 0.984 \\
& & & \\
\textbf{MSE (Validation Data)} & 0.269 & 0.286 & 0.267 \\
\textbf{MSE (Test Data)} & 0.266 & 0.359 & 0.346 \\
\end{longtable}


Table~\ref{tbl-reshpt} presents the network architectures and
hyperparameters of the \glspl{nn} obtained through the different tuning
strategies. The number of units in the first hidden layer, and
consequently in all subsequent layers as described in
Section~\ref{sec-setup}, varies across the approaches. The loss based
tuning strategy selects 256 units in the first hidden layer, while the
weighted and desirability based multi objective approaches identify 1024
and 32 units, respectively, as optimal. In this experiment, no clear
pattern emerges that links the number of hidden units to either
performance or explainability outcomes.

Examining the number of training epochs, the loss based approach
converges at a considerably higher value (512 epochs) compared to the
multi objective strategies, which terminate at 64 and 128 epochs. The
batch size and activation function remain consistent across all three
tuning strategies.

In terms of optimization algorithms, the loss based approach favors
RMSprop, whereas both multi objective tuning runs, which incorporate
\gls{xai} consistency, prefer Adamax. The dropout probability is similar
for the loss based and the weighted multi objective strategies (0.347
and 0.310, respectively), but it is notably lower for the desirability
based approach (0.177).

The learning rate multiplier is identical (0.1) for the loss based and
the weighted multi objective strategies, while it is higher for the
desirability based tuning method.

\begin{longtable}[]{@{}
  >{\raggedright\arraybackslash}p{(\columnwidth - 6\tabcolsep) * \real{0.2727}}
  >{\raggedright\arraybackslash}p{(\columnwidth - 6\tabcolsep) * \real{0.2424}}
  >{\raggedright\arraybackslash}p{(\columnwidth - 6\tabcolsep) * \real{0.2424}}
  >{\raggedright\arraybackslash}p{(\columnwidth - 6\tabcolsep) * \real{0.2424}}@{}}
\caption{Best hyperparameters identified through the three tuning
experiments: single-objective loss-based, multi-objective weighted, and
multi-objective desirability-based.}\label{tbl-reshpt}\tabularnewline
\toprule\noalign{}
\begin{minipage}[b]{\linewidth}\raggedright
Hyperparameter
\end{minipage} & \begin{minipage}[b]{\linewidth}\raggedright
Loss-Based
\end{minipage} & \begin{minipage}[b]{\linewidth}\raggedright
Weighted
\end{minipage} & \begin{minipage}[b]{\linewidth}\raggedright
Desirability
\end{minipage} \\
\midrule\noalign{}
\endfirsthead
\toprule\noalign{}
\begin{minipage}[b]{\linewidth}\raggedright
Hyperparameter
\end{minipage} & \begin{minipage}[b]{\linewidth}\raggedright
Loss-Based
\end{minipage} & \begin{minipage}[b]{\linewidth}\raggedright
Weighted
\end{minipage} & \begin{minipage}[b]{\linewidth}\raggedright
Desirability
\end{minipage} \\
\midrule\noalign{}
\endhead
\bottomrule\noalign{}
\endlastfoot
\textbf{l1} & 256 & 1024 & 32 \\
\textbf{epochs} & 512 & 64 & 128 \\
\textbf{batch size} & 128 & 128 & 128 \\
\textbf{activation function} & \texttt{Swish} & \texttt{Swish} &
\texttt{Swish} \\
\textbf{optimizer} & \texttt{RMSprop} & \texttt{Adamax} &
\texttt{Adamax} \\
\textbf{dropout probability} & 0.347 & 0.310 & 0.177 \\
\textbf{learning rate multiplier} & 0.1 & 0.1 & 1.668 \\
\end{longtable}

\begin{figure}

\centering{

\centering{

\includegraphics[width=1\textwidth,height=\textheight]{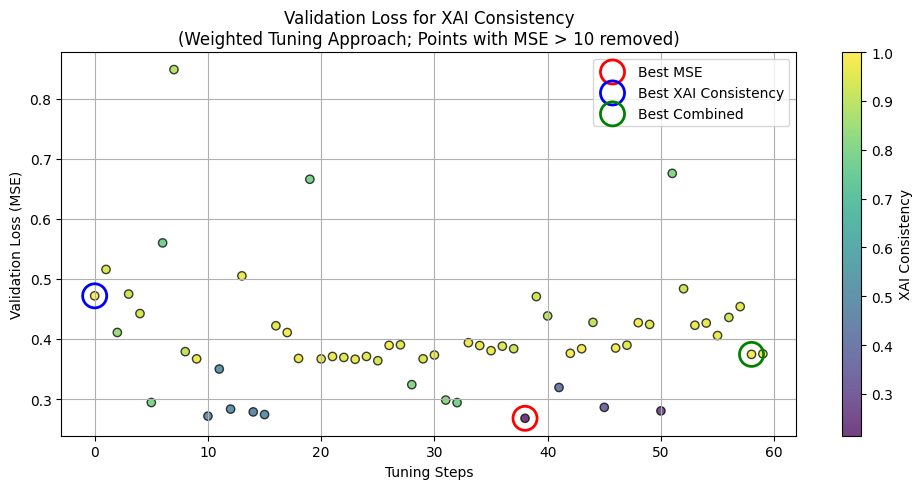}

}

\subcaption{\label{fig-w-tuning}}

\centering{

\includegraphics[width=1\textwidth,height=\textheight]{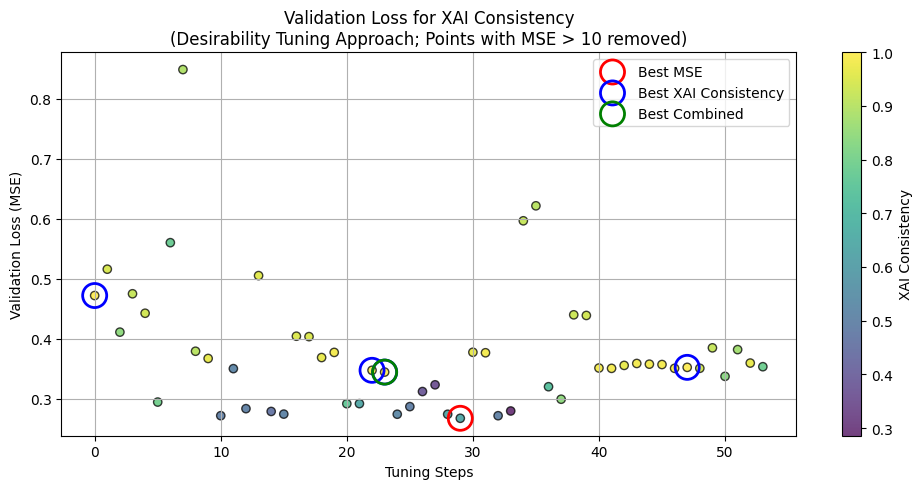}

}

\subcaption{\label{fig-d-tuning}}

}

\caption{\label{fig-tuning}Evolution of performance loss and XAI
consistency during the multi-objective hyperparameter tuning processes.
The top plot (a) shows results for the weighted approach, and the bottom
plot (b) shows results for the desirability-based approach. Colors
indicate the value of XAI consistency (measured by Spearman rank
correlation) across all feature attribution methods. The y-axis
represents the MSE, while the x-axis denotes the tuning steps. Note that
the first 20 points correspond to the initial design prior to the start
of actual tuning. Configurations resulting in an MSE greater than 10
have been excluded from the visualization.}

\end{figure}

The evolution of the tuning process, specifically the search for an
optimal trade-off between \gls{xai} consistency and performance loss for
both multi-objective tuning approaches, is visualized in Figure
\ref{fig-tuning}. The color of each point represents the \gls{xai}
consistency, measured using the Spearman rank correlation.
Configurations shown in yellow correspond to the highest consistency
(100\%), while purple points indicate the lowest consistency values,
around 30\%. The y-axis represents the validation \gls{mse}. It can be
observed that, for both tuning approaches, the configurations yielding
the lowest \gls{mse} values (highlighted with red circles) exhibit
relatively low \gls{xai} consistency (ranging from 30\% to 80\%). In
contrast, the yellow points---indicating configurations with high
\gls{xai} consistency---are clustered around a validation \gls{mse} of
approximately 0.4 for both tuning setups. The configurations with the
highest \gls{xai} consistency are marked with blue circles. The optimal
trade-off, and thus the best overall solution identified by each tuning
strategy, is marked with a green circle. These solutions lie between the
configurations with the best performance and those with the highest
\gls{xai} consistency, representing a balanced compromise.

Overall, these plots reveal three distinct zones within the architecture
configuration space. The first zone contains configurations with both
poor predictive performance and comparatively low \gls{xai} consistency.
The second zone consists of architectures that achieve strong model
performance but suffer from low interpretability. The third zone
(situated between the other two) represents a region, where
configurations maintain high \gls{xai} consistency while still
delivering reasonable performance. This middle zone is particularly
promising for applications that demand both accuracy and explainability.

\section{Summary and Discussion}\label{sec-sum}

In this work, we introduced a framework and a set of metrics to
incorporate \gls{xai} consistency into \gls{mohpo}. We compared two
approaches---equally weighted aggregation and desirability-based
optimization---for balancing \gls{xai} consistency, measured via
Spearman rank correlation, against performance loss, measured as
\gls{mse}, using a demonstrative regression task.

Our results show that the best-performing architectures do not
necessarily achieve the highest \gls{xai} consistency. This suggests
that strong predictive performance does not automatically correspond to
transparent or analyzable decision-making processes. This finding is
particularly relevant in safety- or security-critical applications of
AI, where interpretability is a key requirement.

Building on this observation, we identified three distinct regions
within the architecture configuration space:

\begin{enumerate}
\def\labelenumi{\arabic{enumi}.}
\item
  A region where models perform poorly and also exhibit low \gls{xai}
  consistency,
\item
  A region where models achieve excellent performance but suffer from
  weak interpretability, and
\item
  An intermediate trade-off zone where both objectives---performance and
  consistency---are reasonably balanced.
\end{enumerate}

The third region, in particular, is noteworthy. Configurations in this
zone display high \gls{xai} consistency while maintaining competitive
predictive performance. We hypothesize that models in this trade-off
zone may generalize better and be more robust to out-of-distribution
data, as their decision-making processes are likely to be more stable,
less overly complex, and less prone to overfitting. This assumption
stems from the idea that consistent attributions may indicate simpler,
more generalizable feature usage. However, this hypothesis requires
further empirical validation and will be a focus of future
investigations.

It is important to emphasize that the proposed \gls{xai} consistency
metrics do not assess the correctness of the explanations themselves.
Rather, they quantify the agreement between different feature
attribution methods, and thereby reflect how analyzable or explainable a
model is through standard \gls{xai} techniques. This can be compared to
a linear model: while its coefficients are inherently interpretable,
this interpretability does not necessarily imply a high-quality fit to
the data.

Another noteworthy insight from our experiments is the lack of a clear
correlation between the size or complexity of a \gls{nn} and its
\gls{xai} consistency. Larger or deeper architectures do not inherently
yield more stable explanations. This suggests that explanation
consistency is not trivially linked to network capacity, highlighting
the need for further studies involving more diverse and complex
architectures.

\newpage{}

\section*{References}\label{references}
\addcontentsline{toc}{section}{References}

\phantomsection\label{refs}

\end{document}